\documentclass[conference]{IEEEtran}
\pdfoutput=1  

\IEEEoverridecommandlockouts
\usepackage{cite}
\usepackage{amsmath,amssymb,amsfonts}
\usepackage{algorithmic}
\usepackage{graphicx}
\usepackage{textcomp}
\usepackage{xcolor}
\def\BibTeX{{\rm B\kern-.05em{\sc i\kern-.025em b}\kern-.08em
    T\kern-.1667em\lower.7ex\hbox{E}\kern-.125emX}}
\usepackage[utf8]{inputenc} 
\usepackage[T1]{fontenc}    
\usepackage{hyperref}       
\usepackage{multirow}
\usepackage{url}            
\usepackage{booktabs}       
\usepackage{nicefrac}       
\usepackage{microtype}      
\usepackage{subcaption}

\begin{document}
\title{Neural Operator based Multi-Field Reconstruction of Inner Solar Boundary State}

\author{
\IEEEauthorblockN{
Vignesh Kumar Pandian Sathia\IEEEauthorrefmark{1},
Reza Mansouri\IEEEauthorrefmark{1},
Dustin J. Kempton\IEEEauthorrefmark{1},
Pete Riley\IEEEauthorrefmark{2},
Rafal A. Angryk\IEEEauthorrefmark{1}
}

\IEEEauthorblockA{
\IEEEauthorrefmark{1}\textit{Department of Computer Science, Georgia State University}\\
Atlanta, USA\\
\{spandian1, rmansouri1, dkempton1, rangryk\}@gsu.edu 
}

\IEEEauthorblockA{
\IEEEauthorrefmark{2}\textit{Predictive Sciences Inc.}\\
San Diego, USA\\
pete@predsci.com
}
}

\maketitle

\begin{abstract}
The Solar wind is a continuous flow of charged particles emanating from the solar surface and governed by complex, interacting magnetohydrodynamic processes. Accurate specification of inner-boundary conditions is essential for heliospheric modeling and solar-wind prediction. In many practical applications, only a subset of interacting multi-field variables is directly available, but for a comprehensive view of solar wind prediction and downstream magnetohydrodynamic simulations, a more complete boundary state is required. In this work, we study the problem of learning the multi-field multi-scale solar magnetohydrodynamic state at 30 solar radii ($R_\odot$) using operator learning. Specifically, given the radial velocity and radial magnetic field, we aim to reconstruct the non-radial velocity and magnetic field components, radial and non-radial current density, thermodynamic density, and pressure components. This mapping is highly nonlinear, spatially coupled, and multi-scale, making it a challenging task for data-driven scientific machine learning.

To address this problem, we employ a Local Neural Operator (LocalNO) that learns mappings between input and output function spaces while retaining locality and resolution-awareness. Unlike conventional regression models and autoencoder models, neural operators are better suited for learning structured field-to-field transformations arising from physical systems. The resulting predictions along with inputs are intended to serve as boundary condition variables for future inner-heliospheric modeling pipelines.

\end{abstract}
\section{Introduction}
\label{sec: introduction}

The solar wind is a stream of charged particles emanating from the solar corona and propagating through the inner heliosphere (up to 5 AU) and outer heliosphere toward the heliopause. Understanding and predicting the solar wind is a fundamental problem in space physics and heliophysics because it governs the transport of solar plasma and magnetic fields through the heliosphere, drives space weather disturbances, and directly impacts planetary magnetospheres, satellite operations, communication systems, and human space exploration. Large-scale heliospheric dynamics are governed by coupled magnetohydrodynamic and plasma transport processes, where flow variables, magnetic fields, density, and pressure evolve jointly across radial distance and time. In practice, however, not all quantities required by these models are directly available, and recovering the full physical state from partial observations remains a challenging inverse problem.

In this work, we focus on the solar magnetohydrodynamic state at 30 $R_\odot$, an inner boundary that is particularly relevant for heliospheric simulations. We assume that two physically meaningful quantities are available on this spherical surface: radial velocity and radial magnetic field. From these inputs, we seek to reconstruct the remaining boundary variables needed required for downstream heliospheric modeling, namely velocity in the $\theta$ and $\phi$ directions, magnetic field in the $\theta$ and $\phi$ directions, current density in the radial ($r$), $\theta$, and $\phi$ directions, as well as plasma density $\rho$ and pressure $p$. This is a challenging task for several reasons. First, the target variables are strongly coupled through nonlinear physical structure. Second, the mapping is multi-field and high-dimensional, with each prediction corresponding to a structured field over the entire boundary surface rather than a single scalar. Third, the problem exhibits multi-scale spatial dependence.

Traditional supervised learning methods are not ideally matched to this setting. Convolutional architectures, such as CNNs, are effective at local image-like prediction tasks, but they may be sensitive to discretizations or resolution changes and may struggle to capture the spherical geometry and longitudinal wraparound effectively\cite{reza2025a, reza2025b}. Multilayer Perceptrons may be even more limited, since they are not translation invariant, and may even struggle to capture long-range interactions across the domain. More broadly, physical field reconstruction on a spherical boundary is not merely a pixel-to-pixel regression problem; it is an operator-learning problem in which the model must infer a functional transformation from partial boundary information as input to a complete multi-variable state.

Neural operators provide a natural framework for this class of problems. Rather than learning a finite-dimensional mapping tied to a specific discretization, neural operators aim to approximate operators between infinite-dimensional function spaces. This makes them particularly attractive for scientific machine learning, where the input and output are often fields governed by hidden physical processes. Among these methods, Local Neural Operators offer an appealing balance between expressive power and inductive bias. By emphasizing localized interactions while still learning function-space mappings, they are well suited for settings where the target variables depend on local structure but also emerge from broader physical coupling. For solar magnetohydrodynamic state reconstruction, this locality-aware formulation can help exploit the spatial organization present on the spherical boundary without reducing the task to purely pointwise estimation.

This paper formulates solar boundary-state reconstruction as a multi-output operator-learning problem and evaluates Local Neural Operators for this task at 30 $R_\odot$. Our central premise is that radial velocity and radial magnetic field contain sufficient latent information to reconstruct the missing boundary fields. This problem is closely aligned with the way operational solar-wind prediction systems are constructed, like in WSA/ENLIL \cite{arge2000, odstrcil2003modeling}, the near-Sun solar-wind state is specified from available magnetic-field information and then supplied as an inner-boundary condition to a heliospheric MHD model. Beyond the immediate predictive task, this learned reconstruction can then serve as a complete boundary condition for downstream heliospheric models, providing a holistic surrogate modeling framework.

The contributions of this work are threefold. First, we formulate solar boundary reconstruction as a multi-output operator learning problem, where two observed radial components are used to predict a nine-channel magnetohydrodynamic target. Second, we focus on Local Neural Operators because they combine function-space learning with a locality bias suited to structured boundary fields. Third, the ablation studies explore the effectiveness and feasibility of various components that best suit the current problem setting of spherical boundary reconstruction. While the present work focuses on learning the boundary mapping itself, it is intended as a foundation for future integration with full inner heliospheric simulation frameworks.

The rest of the paper is organized as follows. Section~\ref{sec: related_work} reviews prior work in solar wind prediction, scientific surrogate models, and neural operator methods. Section~\ref{sec: data_preparation} explores the data transformation and preprocessing steps needed to handle the multi-scale components. Section~\ref{sec: methodology} discusses the problem setting, model architecture and training configuration, Section ~\ref{sec:experiments} describes the various experimental setup, evaluation strategy, baseline architectures, and proposed model, Section~\ref{sec: ablation} presents ablation studies used to identify suitable configurations for the proposed reconstruction setting. Section ~\ref{sec: discussion} compares the results and their applicability in this study. To further the scientific research in this domain and share our findings, the code for presented work is available at \cite{sathia2026solarwindprediction}.

\section{Related Work}
\label{sec: related_work}

\subsection{Data-Driven Modeling in Solar and Space Physics}

Physical systems are modeled using two broad paradigms: simulations and inversions. Forward simulations deals with known or assumed physical governing laws and evolving the system forward through numerical solvers or partial differential equations. Models in this paradigm include ballistic extrapolation methods \cite{Schwenn1990, Neugebauer1998}, Arge-Pizzo kinematic models \cite{arge2000, arge2003}, HUX \cite{riley2021hux, riley2022hux}, MAS \cite{riley2001mas}, ENLIL \cite{enlil2023}, and other MHD-based heliospheric models. MAS stands for Magnetohydrodynamic Algorithm outside a Sphere and provides a physics based MHD model to simulate the solar corona by solving for the time-dependent resistive thermodynamic MHD equations in 3D spherical coordinates. It provides the training data used in this study because it contains the target variables described in Section ~\ref{sec: data_preparation}.

The latter paradigm, inversion, aims to infer guiding parameters of the system, from observed quantities. Inversion techniques have become quite useful for surrogate modeling, owing to the computational complexity, time complexity, and time sensitive nature of the problem of the simulation paradigm. Machine learning has become increasingly important for inversion modeling of solar physics and space weather forecasting, where complex nonlinear dynamics, expensive numerical simulations, and limited direct observations motivate surrogate and hybrid modeling approaches. Several inversion techniques have been employed for reconstructing fields from partial states, like Physics-informed Neural Networks \cite{raissi2019physics,cuomo2022scientific, cai2021physics, shaier:pinn}, Physics-encoded Neural Networks \cite{rao:2021:penn1, guo:2024:penn2, qin:2023:penn3}, Physics-guided Neural Networks \cite{pawar:2021:pgml, robinson:2022:pgml2, faroughi:2023:pgml3, faroughi:2024:pgml4}, all of which deal with coordinate grids as inputs and predict target variables. Another framework for the problem setting is directly learning the reconstruction maps from input fields using approaches like CNN-based architectures in \cite{de629solardenoising, xu2020solar, reza2025dkist}, used for solar image denoising using CNNs, Autoencoder-based approaches \cite{cinelli2021vae} and heliophysics foundation models such as Surya \cite{roy2025suryafoundationmodelheliophysics} provide additional examples of representation learning for solar data, and U-Net-based architecture have also been used for detection and analysis of solar eruptive events \cite{stepenyuk2026segmentation} and \cite{reza2025dkist} maps granular structures through semantic segmentation for a potential avenue for linking solar dynamics to surrogate models. In contrast to pixel-wise regression, neural operators can learn field-to-field mappings between functions, as discussed in Section ~\ref{subsec:related_neural_operators}. 

Much of this literature focuses on forecasting one or a small number of scalar or field-valued observables from historical measurements or simulation outputs. In contrast, the present problem concerns reconstruction of an entire multi-field boundary state from only two observed components on a spherical surface. This distinction is important. Solar wind prediction tasks are often framed as time-series forecasting or regression from synoptic maps to scalar outputs at downstream locations. Our setting instead requires estimating missing physical variables that are distributed across the entire spherical boundary and are coupled through underlying hydrodynamic and magnetic structure. As a result, the task is closer to partial state reconstruction than conventional forecasting. This places it at the intersection of space physics, inverse problems, and structured scientific machine learning.

\subsection{Surrogate Modeling for Physical Systems}

Surrogate models have long been used to accelerate numerical simulation and approximate expensive forward models in physics. Early approaches relied on reduced-order models \cite{frangos2010surrogate, benner2015survey}, Gaussian processes \cite{gramacy2020surrogates}, or polynomial approximations. A major challenge in surrogate modeling for physical systems is preserving the physical and geometrical structure of the underlying problem. Many physical quantities are not independent; they arise from coupled governing equations and conservation constraints. As a result, purely black-box regression may achieve low average error while still producing physically inconsistent fields. For boundary reconstruction problems such as ours, this issue is especially relevant because the predicted variables are intended to serve as inputs to downstream inner-heliospheric models.

\subsection{Neural Operators}
\label{subsec:related_neural_operators}

Neural operators were introduced to learn mappings between function spaces rather than between fixed-length vectors. This formulation makes them well suited for parametric PDE problems, where the goal is to map an input field, coefficient, or boundary condition to a solution field. A key advantage of operator learning is that it more naturally reflects the mathematical structure of physical modeling: both the inputs and outputs are functions defined on a domain. Compared with standard neural networks, operator learners can better capture global dependencies and may generalize better across discretizations.

Several neural operator architectures have been proposed in recent years. Among the most prominent are the Fourier Neural Operator (FNO) \cite{li2020fno, du2024glno}, which uses spectral convolution in Fourier space to capture global interactions efficiently, and variants that adapt this idea to different geometries and domains like Spherical Fourier Neural Operator (SFNO) \cite{bonev2023sfno}, Tensorized Fourier Neural Operator (TFNO) \cite{kossaifi2023tfno}, and Geometry-informed Neural Operators \cite{li2023gino}. These methods have demonstrated strong performance on a wide range of PDE benchmarks, including fluid flow, thermodynamic equations, and other spatiotemporal systems. However, purely global spectral methods may not always be ideal for problems where local structure is especially important or where the domain geometry introduces anisotropy and localized physical interactions.

Local Neural Operators (LocalNO) \cite{liu2024lno} was proposed to address some of these limitations by emphasizing localized kernel interactions using differential and integral kernels, while still retaining the operator learning perspective. Local operator formulations learn neighborhood-based interactions that can better capture fine-scale structure, sharp gradient regions, and localized correlations. This makes LocalNO a compelling candidate for solar boundary reconstruction.

\subsection{Operator Learning for Multi-Field and Partial-State Reconstruction}

Most benchmark studies in neural operators consider mappings from one coefficient or forcing field to one solution field. In many scientific applications, however, the task is inherently multi-field: several coupled output variables must be predicted jointly. Joint prediction is often preferable to training separate models because the outputs are physically related and may share latent structure. In our case, the target variables include transverse velocity, transverse magnetic field, current density components, density, and pressure, all of which arise from the same underlying solar plasma state. A multi-output operator learner can exploit these dependencies and potentially produce more consistent reconstructions.

Another closely related theme is partial-state reconstruction, where only a subset of the full system variables is observed and the remainder must be inferred. This setting appears in data assimilation, inverse problems, and reduced sensing scenarios. Although not always framed explicitly as operator learning, many such tasks involve recovering hidden field variables from incomplete measurements. Our formulation can be understood in exactly this way: radial velocity and radial magnetic field are treated as partial observations of the boundary state, and the model learns an operator that reconstructs the missing components.

The present study differs from existing literature in three main ways. First, it focuses on heliospheric boundary reconstruction rather than direct forecasting or full forward simulation. Second, it formulates the task as a multi-field operator learning problem from sparse observed boundary variables to a richer magnetohydrodynamic state. Third, it investigates the use of a Local Neural Operator for this setting, motivated by the need to capture both field-to-field correlation and localized as well as long-range spatial dependence.

\section{Data Preparation}
\label{sec: data_preparation}

Our dataset contains 616 Carrington rotations between 19 February 1975 and 21 December 2020, covering more than four complete solar cycles. The data come from MAS simulations that use boundary conditions at 30 $R_\odot$, derived from observations made by three instruments: the Kitt Peak Observatory (KPO), the SOHO/Michelson Doppler Imager (MDI), and the Solar Dynamics Observatory/Helioseismic and Magnetic Imager (HMI). The dataset contains a total of 907 samples from different instruments and Carrington rotations. Each instance consists of 11 solar wind components on the spherical surface grid ($\theta$, $\phi$) with a ($109\times128$) resolution. These components are the velocity field ($v$), magnetic field ($\mathbf{B}$), current density ($\mathbf{J}$) in three spherical directions ($r, \theta, \phi$) and scalar variables mass density ($\rho$) and the thermal pressure ($p$).

\begin{figure*}
    \centering
    \includegraphics[width=\linewidth]{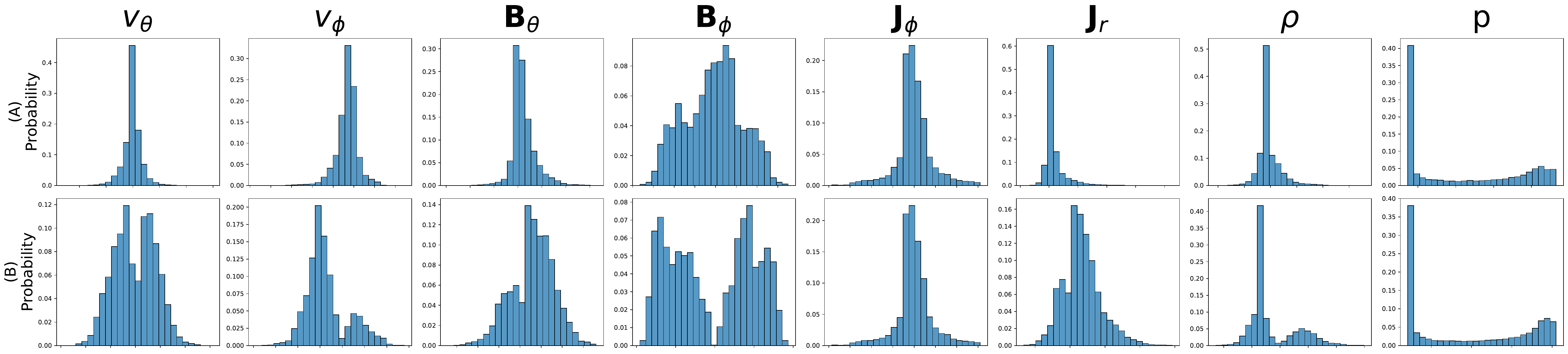}
    \caption{Data distribution. Top row (A) represents the original data, bottom row (B) represents the transformed data with signed square root transformation for variables $v_\theta, v_\phi, \mathbf{B}_\theta, \mathbf{B}_\phi, \mathbf{J}_r, \mathbf{J}_\phi, \rho$, and power transform (power $=1/4$) for pressure $p$. No transformation is needed for $\mathbf{J}_\theta$.}
    \label{fig:data_distribution}
\end{figure*}

The target variables are distributed in different physical ranges, suggesting the need for careful data preprocessing and transformation steps for training. Training the models without these transformations may lead to weaker correlation between the input and target variables as discussed in Section \ref{subsec: ablation_transformation}. To address this issue, we apply a sign-preserving nonlinear transformation. The initial data distributions and the distributions after transformation is shown in Figure \ref{fig:data_distribution} emphasizing the effect of data transformation.

In particular, for variables whose values can be both positive and negative and exhibit narrow-tailed distributions with magnitude less than 1, we apply a signed square-root transform of the form
\begin{equation}
\tilde{x} = \mathrm{sign}(x)\sqrt{|x|}.
\end{equation}

This transformation is applied to $v_{\theta}, v_{\phi}, \mathbf{B}_{\theta},  \mathbf{B}_{\phi}, \mathbf{J}_r,\ \mathbf{J}_{\phi},$ and $\rho$. This transformation expands the relative representation of small and medium-valued regions. Unlike absolute-value transformation, which removes the sign, the signed transform preserves the sign structure of the original field, which is essential for vector quantities such as transverse and longitudinal velocity, transverse and longitudinal magnetic field, and current density. Preserving sign is especially important in this setting because the directionality of the predicted quantities carries physical meaning and influences consistency across the reconstructed multi-field state. As an additional ablation, we also evaluate a signed logarithmic transformation ($\ln$) as discussed in Section \ref{subsec: ablation_transformation}, where signed log transformation is applied to $v_{\theta}, v_{\phi}, \mathbf{B}_{\theta}, \mathbf{B}_{\phi}, \mathbf{J}_r,$ and $ \mathbf{J}_{\phi}$, no transformation is applied to $\mathbf{J}_\theta$, square root transformation is applied to $\rho$, and power ($1/4$) transformation is applied to $p$. We use a 90/10 random train/test split grouped by Carrington rotation. All instrument-specific MAS products corresponding to the same Carrington rotation are assigned to the same split, preventing leakage across correlated samples and evaluates the model on genuinely unseen solar-rotation conditions. 

After transformation, each output channel is normalized independently using min-max normalization, fitted on training data and transformed on testing data, to align the scale to $[0,1]$. Component-wise normalization ensures that all predicted variables contribute more uniformly to the training objective. The normalization stabilizes optimization across multiscale target variables. The predicted normalized target variables are then de-normalized and an inverse transformation is applied and the metrics are scaled to reconstruct actual physical scales in CGS unit system ($cm/s, G, statA/cm^2, g/cm^3,$ and $ dyn/cm^2$ for $v, \mathbf{B, J}, \rho, $ and $p$ components respectively).

The resulting data pipeline therefore consists of five stages: first, extraction of the relevant boundary fields at 30 $R_\odot$; second, sign-preserving nonlinear transformation of selected variables to improve distributional spread while maintaining directional information; third, per-component normalization to place all target channels on comparable numerical scales; fourth, per-component de-normalization of predicted target variables; fifth, applying inverse transformation and scaling to physical units. This preprocessing pipeline plays an important role in stabilizing optimization and improving the quality of multi-field reconstruction as discussed in Section~\ref{subsec: ablation_transformation}.

\section{Methodology}
\label{sec: methodology}

\subsection{Problem Formulation and Proposed Architecture}

We formulate solar boundary reconstruction as a supervised operator learning problem. Let $u = \left[v_r, \mathbf{B}_r\right]$ denote the observed input fields on the spherical boundary at 30 solar radii, where $v_r$ is the radial velocity and $\mathbf{B}_r$ is the radial magnetic field. The objective is 
to map these partial observations to the full target state:
\begin{equation}
y = \left[v_\theta, v_\phi, \mathbf{B}_\theta, \mathbf{B}_\phi, \mathbf{J}_\theta, \mathbf{J}_\phi, \mathbf{J}_r, \rho, p\right].
\end{equation}

\begin{figure}
    \centering
    \includegraphics[width=\linewidth]{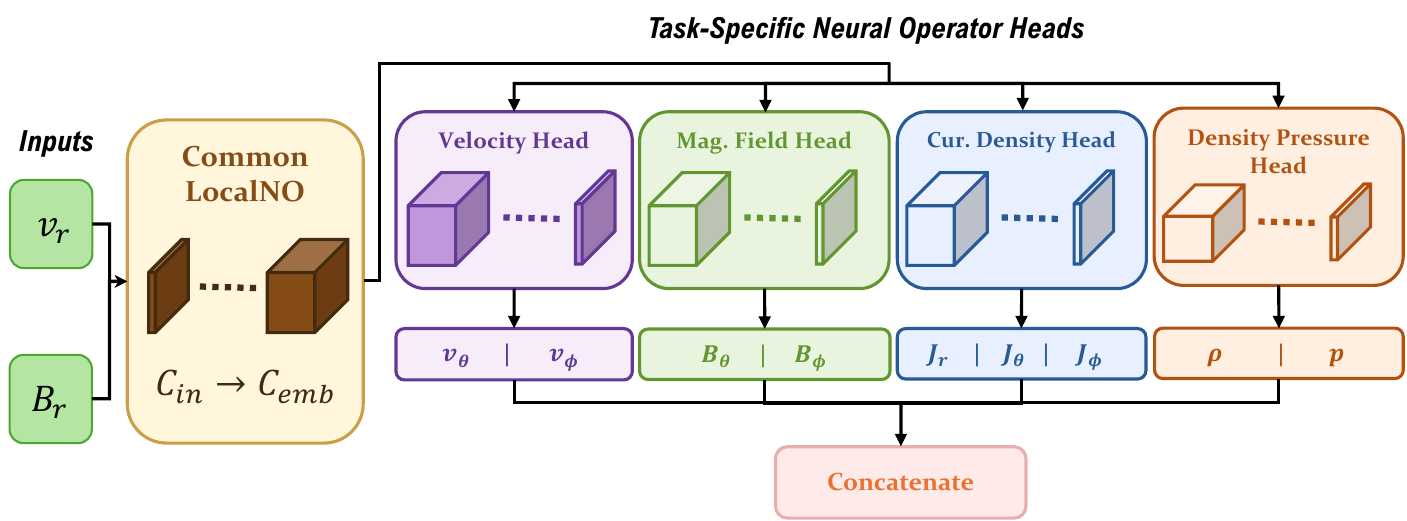}
    \caption{Model Architecture of the proposed methodology employing a common Neural Operator that takes input channels and maps them into latent channels. The output of common NO is fed into task specific heads for correlated variables.}
    \label{fig:model_arch}
\end{figure}
The model architecture is shown in Figure \ref{fig:model_arch}. The learning task is therefore not a scalar regression problem, but a structured field-to-field mapping in which the model must infer missing physical quantities from only a limited subset of observed variables.

\begin{figure*}
    \centering
    \begin{minipage}[t]{0.49\textwidth}
        \centering
        \includegraphics[width=\linewidth]{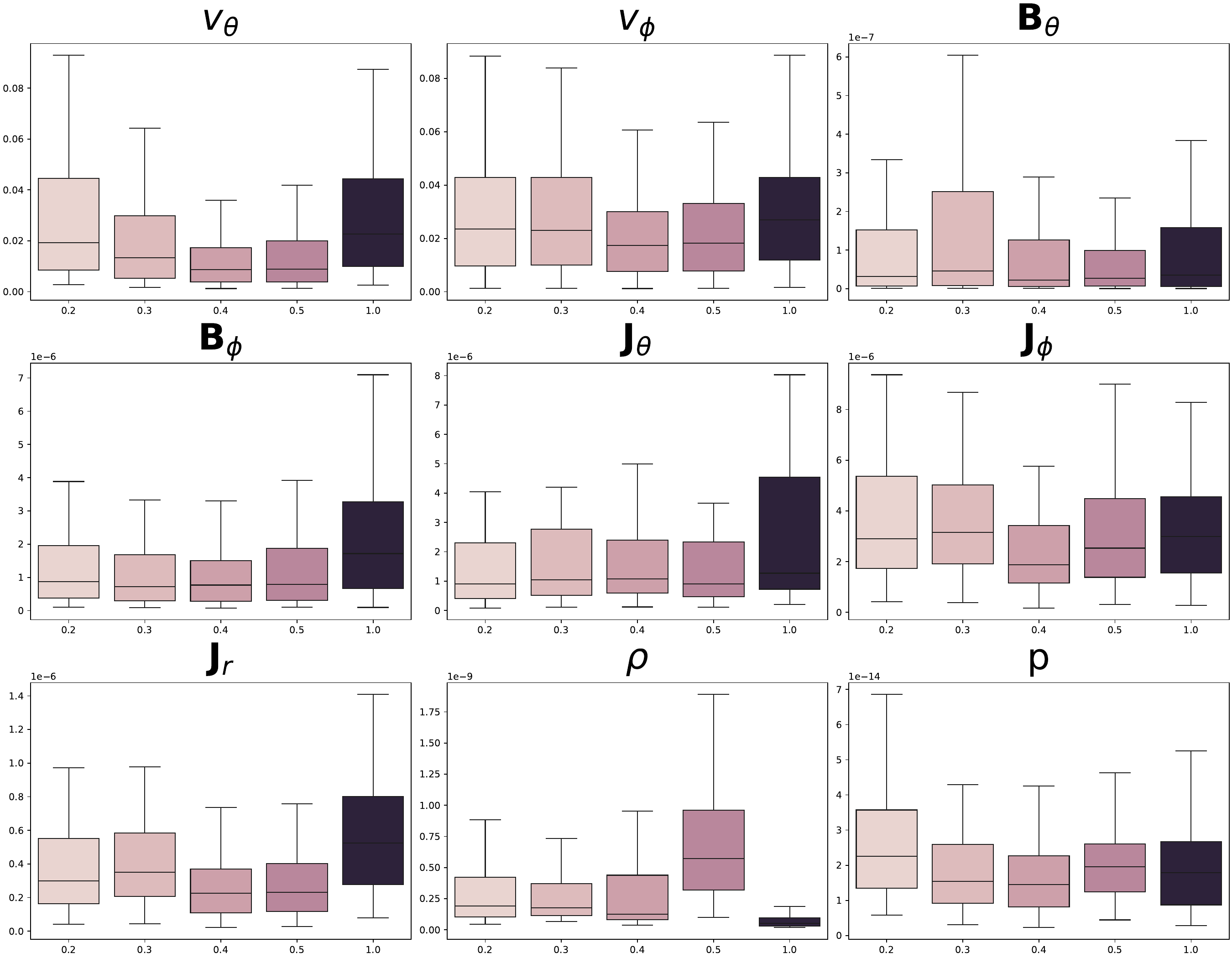}
    \end{minipage}
    \hfill
    \begin{minipage}[t]{0.49\textwidth}
        \centering
        \includegraphics[width=\linewidth]{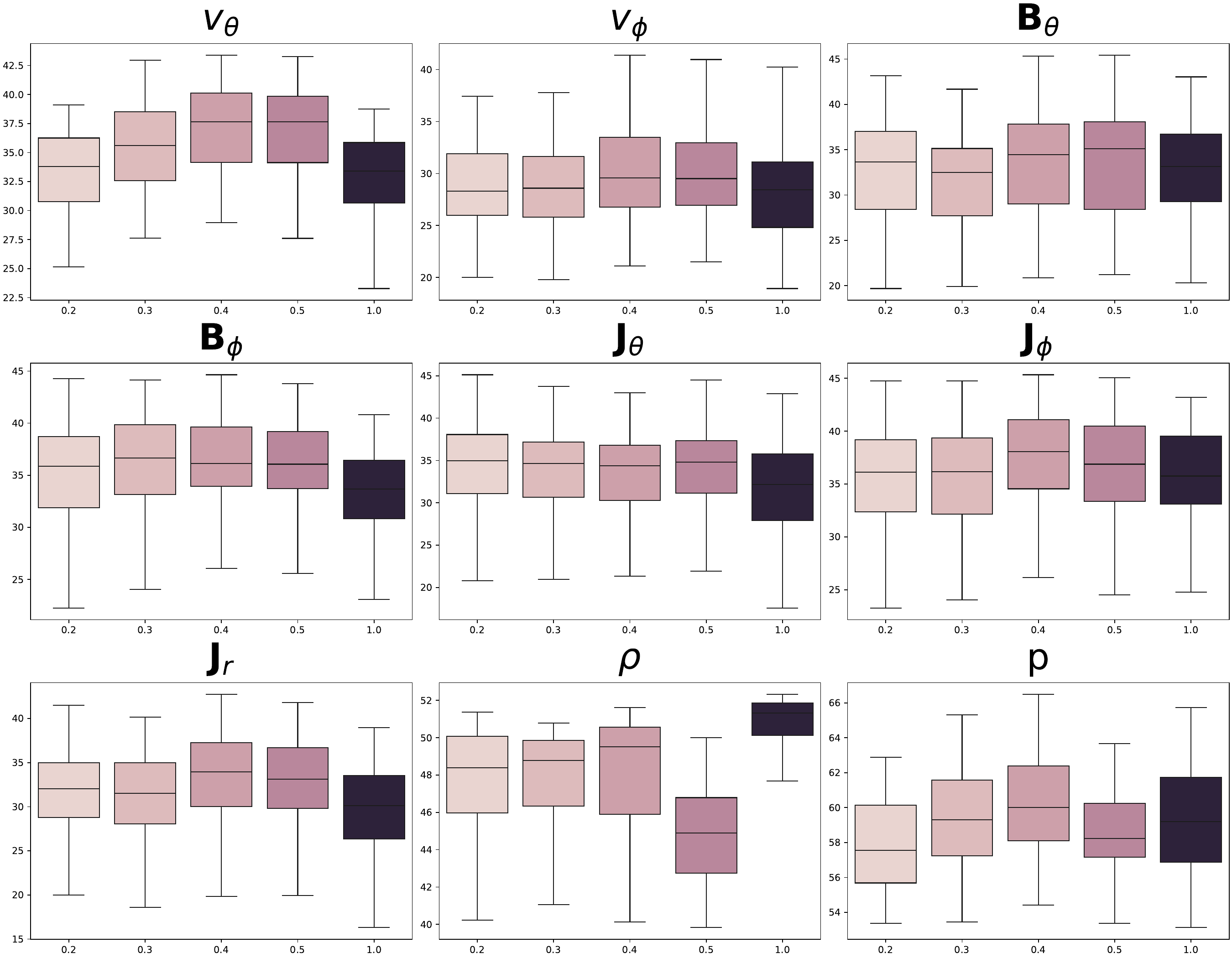}
    \end{minipage}

    \caption{Component-wise distribution of evaluation metrics for rank ablation. Left is the Mean Square Error (MSE) of the 9 components to measure point-wise reconstruction error. Right is the Peak Signal-to-Noise Ratio (PSNR) to measure the signal normalized reconstruction fidelity.}
    \label{fig:ablation_rank}
\end{figure*}
\subsection{Model Families}

To evaluate the suitability of operator learning for this problem, we study three model families: the Spherical Fourier Neural Operator (SFNO), the Tensorized Fourier Neural Operator (TFNO), and the Local Neural Operator (LocalNO).

The SFNO serves as a strong baseline for operator learning in scientific machine learning on spherical domains. It parameterizes the integral kernel in Fourier space and uses spherical convolution to propagate information across the spatial domain. This design allows SFNO to capture long-range interactions efficiently using spherical harmonics and has made it highly successful for learning PDE solution operators of data embedded in spherical space. In our setting, SFNO provides a global operator-learning baseline for mapping radial boundary observations to the missing multi-field magnetohydrodynamic state. The second baseline model for this study is the TFNO architecture that extends the FNO by factorizing the spectral weight tensors, thereby reducing the number of learnable parameters while retaining the overall spectral operator structure. This factorized design makes TFNO attractive when balancing model expressivity and parameter efficiency. 

Our target model class is the LocalNO. Unlike purely global spectral architectures, LocalNO introduces a stronger locality bias by combining localized differential and integral kernels with operator learning in addition to spectral/spherical convolution. This is particularly relevant for solar boundary reconstruction, where the missing target variables may depend strongly on local neighboring structure even though global coupling remains important. We investigate LocalNO variants that use spherical convolution without domain padding, instead of standard spectral convolution, which is designed to better align with the geometry of the solar boundary surface. Since the data live naturally on a spherical manifold rather than a planar Euclidean grid, spherical convolution offers an alternative inductive bias that may better respect the domain structure,  making domain padding and spectral convolution less suitable for this setting.

\section{Experiments}
\label{sec:experiments}

\subsection{Experimental Setup}

We conduct experiments to evaluate the quality of reconstructing the nine-dimensional solar magnetohydrodynamic target state from $v_r$ and $\mathbf{B}_r$ at 30 $R_\odot$. The models compared in this study include SFNO, TFNO, and Local Neural Operator variants with spherical convolution without domain padding. SFNO serves as a global spherical baseline, TFNO as a parameter-efficient tensorized spectral baseline, and LocalNO as the locality-aware operator-learning family. We evaluated LocalNO variants with different convolutional modules during model selection. Because the final model uses the spherical convolution setting, which is better aligned with the angular solar boundary grid, we use this configuration for all reported LocalNO results.

All models are trained on the same preprocessed dataset and evaluated using the same protocol to ensure a fair comparison. All models are trained for 100 epochs using the same hyper-parameters like \texttt{n\_modes = }$(109,128)$, Tucker Factorization, \texttt{batch\_size=}$16$, \texttt{n\_layers=}$4$, and \texttt{lr=}$1e-4$. The focus of the experiments is on field reconstruction quality rather than computational speed alone, since the final objective is to obtain accurate boundary conditions for future heliospheric modeling. Because the outputs are multi-field and spatially structured, evaluation must capture both per-sample fidelity and per-component behavior. All models are trained on \texttt{NVIDIA} A40 GPUs with 48 GB of GPU memory, using \texttt{PyTorch 2.11.0}, and the \texttt{neuraloperator v2.0.0} package for operator models.

\subsection{Evaluation Metrics}

To comprehensively assess reconstruction quality, we report multiple complementary metrics, including Peak Signal-to-Noise Ratio (PSNR), Universal Quality Index (UQI)~\cite{uiqi}, Earth Mover's Distance (EMD) \cite{emv}, Anomaly Correlation Coefficient (ACC), Normalized Nash-Sutcliffe Efficiency (NNSE) \cite{nnse}, and Mean Squared Error (MSE). These metrics are computed sample-wise and component-wise. The metrics measure the quality of reconstructing an entire example across all spatial locations, while revealing individual component-wise distribution of the predicted variables. This distinction is important because a model may achieve good average performance overall while failing on specific target components. By reporting both views, we obtain a more complete picture of model behavior.

PSNR measures reconstruction fidelity in terms of signal-to-noise ratio and is especially useful for quantifying pixel-level similarity. UQI captures structural similarity and perceptual consistency, which are important when assessing whether predicted fields preserve spatial organization rather than only numerical closeness. MSE measures the component-wise grid-based reconstruction error.

\section{Ablation Study}
\label{sec: ablation}

\subsection{Factorization Rank}
\label{subsec: ablation_rank}

To better understand the role of model capacity in tensorized and local operator learning, we conduct an ablation study over rank values. Specifically, we evaluate ranks $r \in \{0.2, 0.3, 0.4, 0.5, 1.0\}$.

The purpose of this study is to analyze the trade-off between compression and expressivity. Lower ranks impose a stronger bottleneck on the operator representation, which improves parameter efficiency and regularization but can limit the model's ability to capture the complex coupling among solar hydrodynamic fields. Higher ranks provide greater expressive capacity, but they may reduce the benefits of structured factorization leading to diminished returns or over-parameterization.

Our experiments show that rank $0.4$ yields the best overall trade-off across the evaluated metrics. This suggests that the solar boundary reconstruction problem, as shown in Figure \ref{fig:ablation_rank} and Table \ref{tab:ablation_rank}, benefits from a moderate amount of factorized capacity: too little rank underfits the nonlinear field relationships, while too much rank does not translate into corresponding gains in reconstruction quality. The result is particularly important because it indicates the need to balance representational flexibility with inductive regularization rather than simply maximizing capacity.

\begin{table*}[htbp]
\centering
\caption{Comparison of model performance when trained across factorization rank when using  Tucker factorization. The metric values are aggregated over the components. For MSE and EMD~\cite{emv} metrics, lower values are better, and for ACC, NNSE~\cite{nnse}, PSNR, and UQI, higher values are better.}
\label{tab:ablation_rank}
\resizebox{\textwidth}{!}{%
\begin{tabular}{|c|cc|cc|cc|cc|cc|cc|}
\hline
\multirow{2}{*}{Rank} 
& \multicolumn{2}{c|}{ACC} 
& \multicolumn{2}{c|}{EMD} 
& \multicolumn{2}{c|}{MSE} 
& \multicolumn{2}{c|}{NNSE} 
& \multicolumn{2}{c|}{PSNR} 
& \multicolumn{2}{c|}{UQI} \\
\cline{2-13}
& Mean & Std. 
& Mean & Std. 
& Mean & Std. 
& Mean & Std. 
& Mean & Std. 
& Mean & Std. \\
\hline
0.2 & 0.9833 & 0.0241 & 0.0097 & 0.0194 & 0.0070 & 0.0192 & 0.9676 & 0.0430 & 37.3505 & 9.9128 & 0.8873 & 0.1108 \\
0.3 & 0.9830 & 0.0256 & 0.0095 & 0.0203 & 0.0058 & 0.0169 & 0.9673 & 0.0445 & 37.5967 & 10.2752 & 0.8836 & 0.1149 \\
0.4 & \textbf{0.9874} & 0.0205 & \textbf{0.0060} & 0.0121 & \textbf{0.0040} & 0.0114 & \textbf{0.9755} & 0.0364 & \textbf{38.6972} & 10.0915 & 0.9071 & 0.0929 \\
0.5 & \textbf{0.9874} & \textbf{0.0203} & 0.0063 & 0.0127 & 0.0043 & 0.0126 & 0.9746 & 0.0381 & 38.0046 & \textbf{9.4680} & \textbf{0.9081} & \textbf{0.0914} \\
1.0 & 0.9787 & 0.0338 & 0.0105 & 0.0215 & 0.0078 & 0.0218 & 0.9602 & 0.0549 & 37.0086 & 11.1103 & 0.8786 & 0.1204 \\
\hline
\end{tabular}%
}
\end{table*}




\subsection{Data Transformation}
\label{subsec: ablation_transformation}

An additional experimental factor is the use of signed square-root transformation  and signed log transformation during preprocessing. Without this transformation, variables with highly concentrated or skewed distributions remain difficult to learn even after normalization, motivating nonlinear transformations that improve distributional spread, making it more difficult for the model to learn finer structure in low- to medium-magnitude regions. The signed square-root transform improves the spread of the data while preserving the sign of physically directional quantities. Empirically, this preprocessing improves stability and leads to better reconstruction across multiple target components as shown in Table ~\ref{tab:ablation_data_transformation}. We also compared the performance across the metrics averaged across components for both square-root and log transformation. This highlights the importance of physically informed preprocessing in multi-field scientific learning problems.

\begin{table*}[htbp]
\centering
\caption{Comparison of model performance for ablation study across data transformation strategies. The metric values are aggregated over the components. For MSE and EMD metrics, lower values are better, and for others higher values are better.}
\label{tab:ablation_data_transformation}
\resizebox{\textwidth}{!}{%
\begin{tabular}{|c|cc|cc|cc|cc|cc|cc|}
\hline
\multirow{2}{*}{Transformation}
& \multicolumn{2}{c|}{ACC}
& \multicolumn{2}{c|}{EMD}
& \multicolumn{2}{c|}{MSE}
& \multicolumn{2}{c|}{NNSE}
& \multicolumn{2}{c|}{PSNR}
& \multicolumn{2}{c|}{UQI} \\
\cline{2-13}
& Mean & Std.
& Mean & Std.
& Mean & Std.
& Mean & Std.
& Mean & Std.
& Mean & Std. \\
\hline
None & 0.9823 & 0.0390 & 0.0072 & 0.0169 & 0.0052 & 0.0173 & 0.9686 & 0.0644 & \textbf{39.6712} & 10.6426 & 0.8607 & 0.1530 \\
Square-root & \textbf{0.9874} & \textbf{0.0205} & \textbf{0.0060} & \textbf{0.0121} & \textbf{0.0040} & \textbf{0.0114} & \textbf{0.9755} & \textbf{0.0364} & 38.6972 & \textbf{10.0915} & \textbf{0.9071} & \textbf{0.0929} \\
Natural Log& 0.9806 & 0.0407 & 0.0081 & 0.0193 & 0.0062 & 0.0207 & 0.9642 & 0.0728 & 39.1683 & 10.7892 & 0.8527 & 0.1603 \\
\hline
\end{tabular}%
}
\end{table*}
\section{Results and Discussion}
\label{sec: discussion}
\begin{figure*}
    \centering
    \begin{subfigure}{\linewidth}
        \centering
        \includegraphics[width=\linewidth]{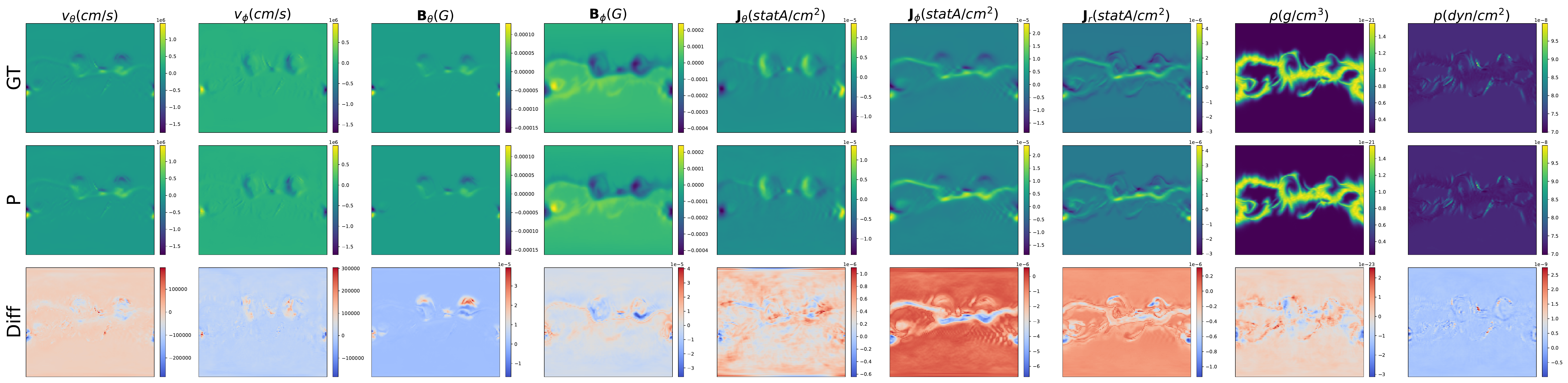}
        \caption{Carrington Rotation 1653 during low solar activity.}
        \label{fig:reconstruction_low_activity}
    \end{subfigure}
    \begin{subfigure}{\linewidth}
        \centering
        \includegraphics[width=\linewidth]{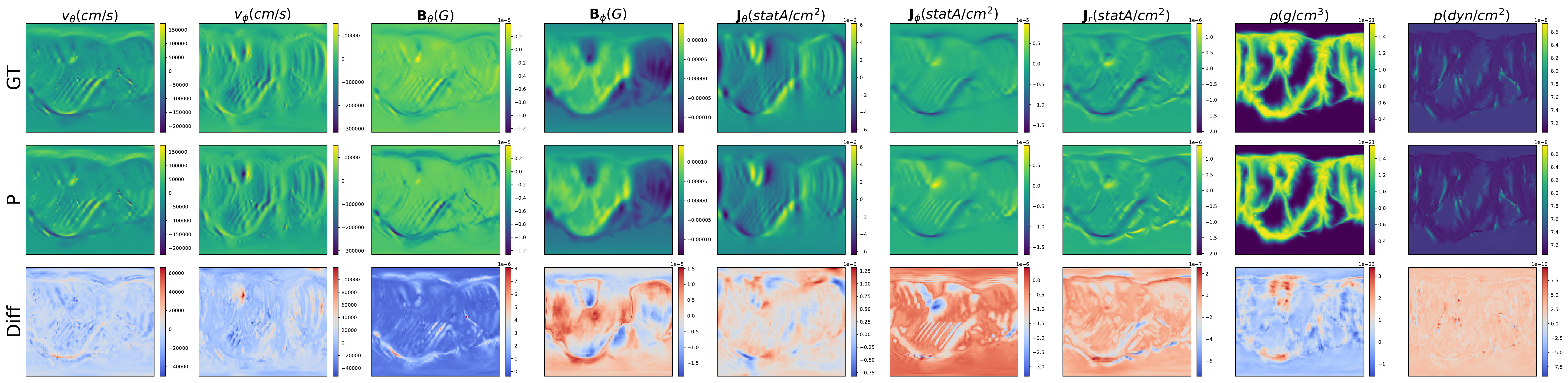}
        \caption{Carrington Rotation 2113 during high solar activity.}
        \label{fig:reconstruction_high_activity}
    \end{subfigure}

    \caption{Component-wise reconstruction of solar MHD state variables. In each subfigure, columns represent the target variables. The top row shows the ground truth (GT), the middle row shows the model prediction (P), and the bottom row shows the residual computed as $GT - P$.}
    \label{fig:reconstruction_low_high_activity}
\end{figure*}

\begin{table*}
\centering
\caption{Comparison of model performance with respect to baseline models. The values are aggregated over components. lower values are preferred for MSE and EMD, higher value is preferred for the remaining metrics.}
\label{tab:model_baseline_comparison}
\resizebox{\textwidth}{!}{%
\begin{tabular}{|c|c|cc|cc|cc|cc|cc|cc|}
\hline
\multirow{2}{*}{Model}
& {Params}
& \multicolumn{2}{c|}{ACC}
& \multicolumn{2}{c|}{EMD}
& \multicolumn{2}{c|}{MSE}
& \multicolumn{2}{c|}{NNSE}
& \multicolumn{2}{c|}{PSNR}
& \multicolumn{2}{c|}{UQI} \\
\cline{3-14}
&(in M)& Mean & Std.
& Mean & Std.
& Mean & Std.
& Mean & Std.
& Mean & Std.
& Mean & Std. \\
\hline
SFNO    &31.65& 0.6307 & 0.4542 & 0.0416 & 0.1106 & 0.0870 & 0.3197 & 0.7651 & 0.2274 & 28.6258 & 11.4363 & 0.5662 & 0.4028 \\
TFNO    &412.92& 0.9488&0.0753&0.0150&0.0337&0.0194&0.0555&0.9200&	0.0937&31.8203&8.4401&0.8035&0.1882 \\
\textbf{LocalNO} &14.82& \textbf{0.9874} & \textbf{0.0205} & \textbf{0.0060} & \textbf{0.0121} & \textbf{0.0040} & \textbf{0.0114} & \textbf{0.9755} & \textbf{0.0364} & \textbf{38.6972} & \textbf{10.0915} & \textbf{0.9071} & \textbf{0.0929} \\
\hline
\end{tabular}%
}
\end{table*}

Table~\ref{tab:model_baseline_comparison} compares the proposed architecture with the SFNO and TFNO baselines. The LocalNO outperforms baselines across reported metrics averaged over each component. The spread is also smaller, suggesting more consistent performance across samples and output components. This supports the hypothesis that locality is a useful inductive bias for reconstructing coupled fields from partial radial observations.

Although SFNO representations are useful for global spherical-domain modeling, they may be less effective when reconstruction depends heavily on localized structures or when the variables have different scales and spatial patterns. The much larger standard deviations for SFNO indicate that its performance is unstable across components. The weaker performance of TFNO compared with LocalNO is consistent with the broader trend that a purely global or tensorized spectral operator may not be sufficient for this inherently spherical domain problem.

Table \ref{tab:ablation_rank} and Figure \ref{fig:ablation_rank} show the effect of factorization rank for the LocalNO. The increase in model performance by increasing rank values from $0.2$ to $0.4$ suggests limited expressivity and the loss of information from strong compression arising from lower rank values. The poorer performance as we increase the rank from $0.5$ up to $1.0$ can be attributed to model over-parameterization and reduction of reconstruction flexibility.

Table \ref{tab:ablation_data_transformation} show the effect of data transformation for more effective training. Signed square root transformation seems to perform better across the metrics over signed log transformation. This could be attributed to higher diverging nature of log transformation between positive and negative values resulting in poorer reconstruction of  small-magnitude field values.

Although this work presents a promising approach, it has several limitations. First, the presented work is only evaluated on  simulations, not real-world data. Second, it does not consider the radiative interactions of the field variables and is beyond the scope of this work. Finally, it does not consider high-resolution simulation due to differences in the spatial resolution and would require additional preprocessing.



\section{Conclusion}

This work studied the problem of reconstructing the solar magnetohydrodynamic boundary state at 30 solar radii from only radial velocity and radial magnetic field. This reconstructed state is intended to serve as a boundary condition for future heliospheric simulations. Our findings indicate that model design choices strongly affect reconstruction quality. In particular, locality-aware operator learning provides a compelling framework for this problem. The rank ablation study further shows that moderate factorization is most effective, with rank $0.4$ achieving the best balance between compression and expressive power. In addition, the signed square-root transformation used during preprocessing improves the spread of the data while preserving physically meaningful sign information, which contributes to more stable and effective learning. 

Overall, these results support the use of operator-learning surrogates for partial boundary state reconstruction in solar physics. In future work, the learned reconstructions can be integrated into downstream heliospheric simulation pipelines, and additional physics-aware constraints may be incorporated to further improve consistency and generalization.

\bibliographystyle{ieeetr}
\bibliography{main}
\end{document}